\documentclass[10pt,final,journal,twocolumn]{IEEEtran}
\usepackage{amsmath}
\usepackage{graphicx, epstopdf}
\usepackage{setspace}
\usepackage{acronym}
\usepackage{commath}
\usepackage{mathrsfs}
\usepackage{mathtools}
\usepackage{calc}
\usepackage[top=1in, left=0.7in, right=0.7in]{geometry}
\usepackage{ifthen}
\usepackage{textcomp}
\usepackage{float}
\usepackage[tight,footnotesize]{subfigure}
\usepackage{cite}
\usepackage{bm}
\usepackage{url}
\usepackage[utf8]{inputenc}
\usepackage{mathtools}
\usepackage{amssymb}
\usepackage{amsthm}
\usepackage{mathrsfs}
\usepackage{bm}
\usepackage{multirow}
\usepackage{multicol}
\usepackage{xspace}
\usepackage{tabularx}

\usepackage{cite}
\usepackage{amsmath,amssymb,amsfonts}
\usepackage{algorithmic}
\usepackage{algorithm}
\usepackage{graphicx}
\usepackage{textcomp}

\usepackage{xcolor}
\def\BibTeX{{\rm B\kern-.05em{\sc i\kern-.025em b}\kern-.08em
    T\kern-.1667em\lower.7ex\hbox{E}\kern-.125emX}}

\begin{document}

\title{Contextualized Autonomous Drone Navigation using LLMs Deployed in Edge-Cloud Computing}

\author{Hongqian Chen$^{1}$, Yun Tang$^{1}$, Antonios Tsourdos$^{1}$, Weisi Guo$^{1*}$
\thanks{$^{1}$All authors are with the School of Aerospace, Transport, and Manufacturing, Cranfield University, Bedford, UK. $^{*}$Corresponding author: weisi.guo@cranfield.ac.uk. 
This work is supported by Leonardo UK and EPSRC TAS-S: Trustworthy Autonomous Systems: Security (EP/V026763/1) and EPSRC CHEDDAR: Communications Hub For Empowering Distributed ClouD Computing Applications And Research (EP/X040518/1) (EP/Y037421/1).}
}

\maketitle

\begin{abstract}
Autonomous navigation is usually trained offline in diverse scenarios and fine-tuned online subject to real-world experiences. However, the real world is dynamic and changeable, and many environmental encounters/effects are not accounted for in real-time due to difficulties in describing them within offline training data or hard to describe even in online scenarios. However, we know that the human operator can describe these dynamic environmental encounters through natural language, adding semantic context. The research is to deploy Large Language Models (LLMs) to perform real-time contextual code adjustment to autonomous navigation. The challenge not evaluated in literature is what LLMs are appropriate and where should these computationally heavy algorithms sit in the computation-communication edge-cloud computing architectures. In this paper, we evaluate how different LLMs can adjust both the navigation map parameters dynamically (e.g., contour map shaping) and also derive navigation task instruction sets. We then evaluate which LLMs are most suitable and where they should sit in future edge-cloud of 6G telecommunication architectures.
\end{abstract}


\section{Introduction}
Smart cities will increasingly integrate autonomous vehicles in the ground and in the air (eVTOLs and air taxis) for a variety of purposes. Autonomous navigation has been a subject of intensive research and development in recent decades, especially in urban and congested areas \cite{9339877, ICUAS}. Dynamically adjusting autonomous navigation in response to emerging situations, effects, and values remains hard encoded. For example, a global path planning algorithm (e.g., A-star) may toggle between a set of penalty functions (e.g., low carbon footprint route, fastest route, smoothest route, avoid certain locations route) based on user requirements. Although these reflect certain user values, the algorithms cannot consider all personal preferences and situational contexts at the offline programming and training stage. The opportunity is that LLMs can enhance navigation performance by personalizing the code to adapt to emerging real-world changes \cite{PMLR21, 10802696}. For example, as shown in Fig.\ref{figs1}, the human user may interpret certain obstacles differently depending on context or situation of that day, this is difficult to pre-code into the algorithm. As such, LLMs can help translate user reactions in real-time into potential fields that augment the navigation algorithm. 

\subsection{Review of LLM-informed Navigation}

The attention-based transformer architecture in LLMs utilizes a self-attention mechanism to process input data. Unlike previous models such as RNNs and LSTMs, which process data sequentially, transformers handle input data in parallel. This parallel processing not only enhances model efficiency but also strengthens the model's ability to understand the context within language. During the training of an LLM, the model typically undergoes two main phases: pre-training and fine-tuning:
\begin{itemize}
    \item In the pre-training phase, the model is trained on a large amount of text using unsupervised learning techniques to gain a general understanding of language and context. 
    \item In the fine-tuning phase, the model is adjusted on a smaller, task-specific dataset to perform well on specific tasks such as answering questions or classification. 
\end{itemize}
LLMs as such have the strong ability to semantically contextualize a wide range of algorithms. For example, in \cite{17} the research explores the potential of utilizing GPT-3.5-turbo from OpenAI, for robotic path planning. In recent years \cite{10518077}, work has been done to use LLMs to generate rare scenarios so that offline training in autonomous navigation can achieve safer and more contextualized path planning and improve trustworthy \cite{10801932}. The path planning element maybe an optimisation route based on globally known parameters (e.g., A-star) and/or adjust locally to emerging challenges using an offline trained data-driven reinforcement learning (RL) algorithm (e.g., toggle policies in RL \cite{SMC} and improve motion control \cite{13}). In our recent work \cite{TAS}, we encoded different philosophical and social values in how path planning algorithms react to emerging hazards and challenges such that users can perform real-time online soft-encoding of algorithms during a navigation process using natural language. More recent advances have also added multi-modal data such as language-image associations to better understand landmarks and obstacles in navigation \cite{15} and interact with them \cite{10610751}.

Most existing work in this emerging area has largely focused on a single navigation process (e.g., route planning around obstacles) and not consider multiple sub-task-based navigation (e.g., take-off, track target, and land). Neither has existing work considered the computation requirements of LLMs in realistic distributed mobile systems and has largely used cloud-based services with high latency and a strong wired communication network requirement.

\subsection{Gap and Novelty}
The research is to deploy Large Language Models (LLMs) to perform real-time contextual code adjustment to autonomous navigation. LLMs translate user natural language into formal code generation, modifications, or parameterizations in real time. What is not sufficient acknowledged are two important areas: (1) how can LLMs generate instruction sets for autonomous navigation that requires several sub-tasks, and (2) which LLMs are most suitable for mobile agents and where they should sit in future edge-fog-cloud continuum of 6G telecommunication architectures. We aim to show how we can create realistic urban drone navigation scenarios via a simulation platform and:
\begin{itemize}
    \item Using LLMs to contextualize path planning - digesting situational contexts and new obstacle data. 
    \item Using Small language models to generate sequence of sub-tasks and tactical control commands for a planned path;
    \item Evaluate the suitability of different LLMs (GPT, LLama, Moonshot, Qwen) in terms of computation and latency and where they should sit architecturally to enable the above functionalities;
\end{itemize}
Combined, they offer a new level of LLM-navigation architecture for real-time navigation situations for future urban mobility. 

\begin{figure}[!t]
\centering
\includegraphics[width=3.5in]{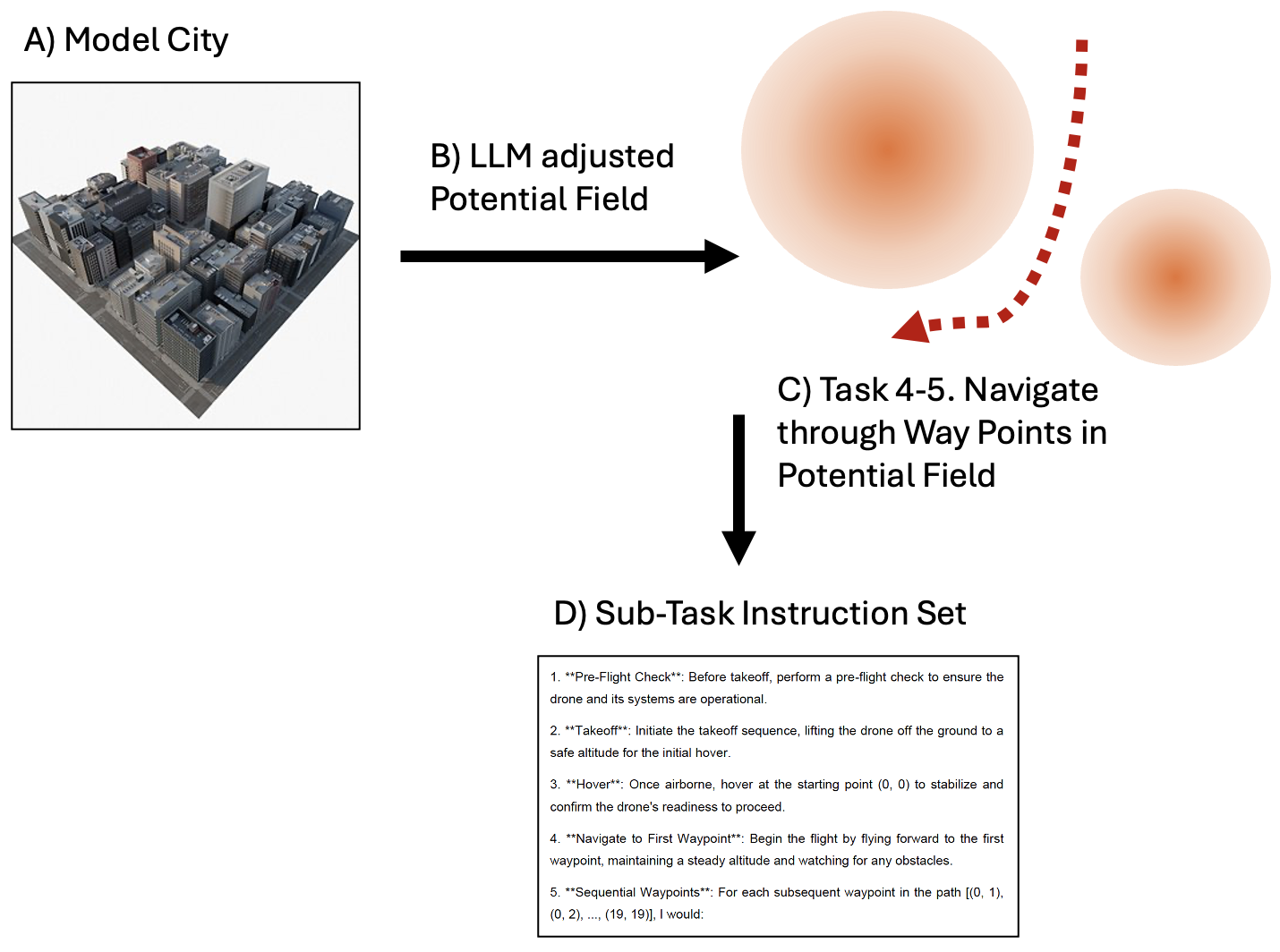}
\caption{System Diagram of LLM-navigation: (A) 3D city map, (B) Mission definition, (C) LLM generated instruction set for sub-tasks, (D) Navigation via way points and potential fields, and (E) Contextualized potential field adjustment using LLMs.}
\label{figs1}
\end{figure}

\section{System Setup and Methodology}

The overall system architecture shown in Fig.\ref{figs1} is formed of several intertwined components that aims at navigating whilst minimizing a cost function related to obstacle and situational potential fields: (A) 3D city map, (B) Contextualized potential field adjustment using LLMs to enable (C) Path planning via way points and potential fields, and LLM generated instruction set for drone flight sub-tasks for the path planned. 

\begin{figure}[!t]
\centering
\includegraphics[width=3.3in]{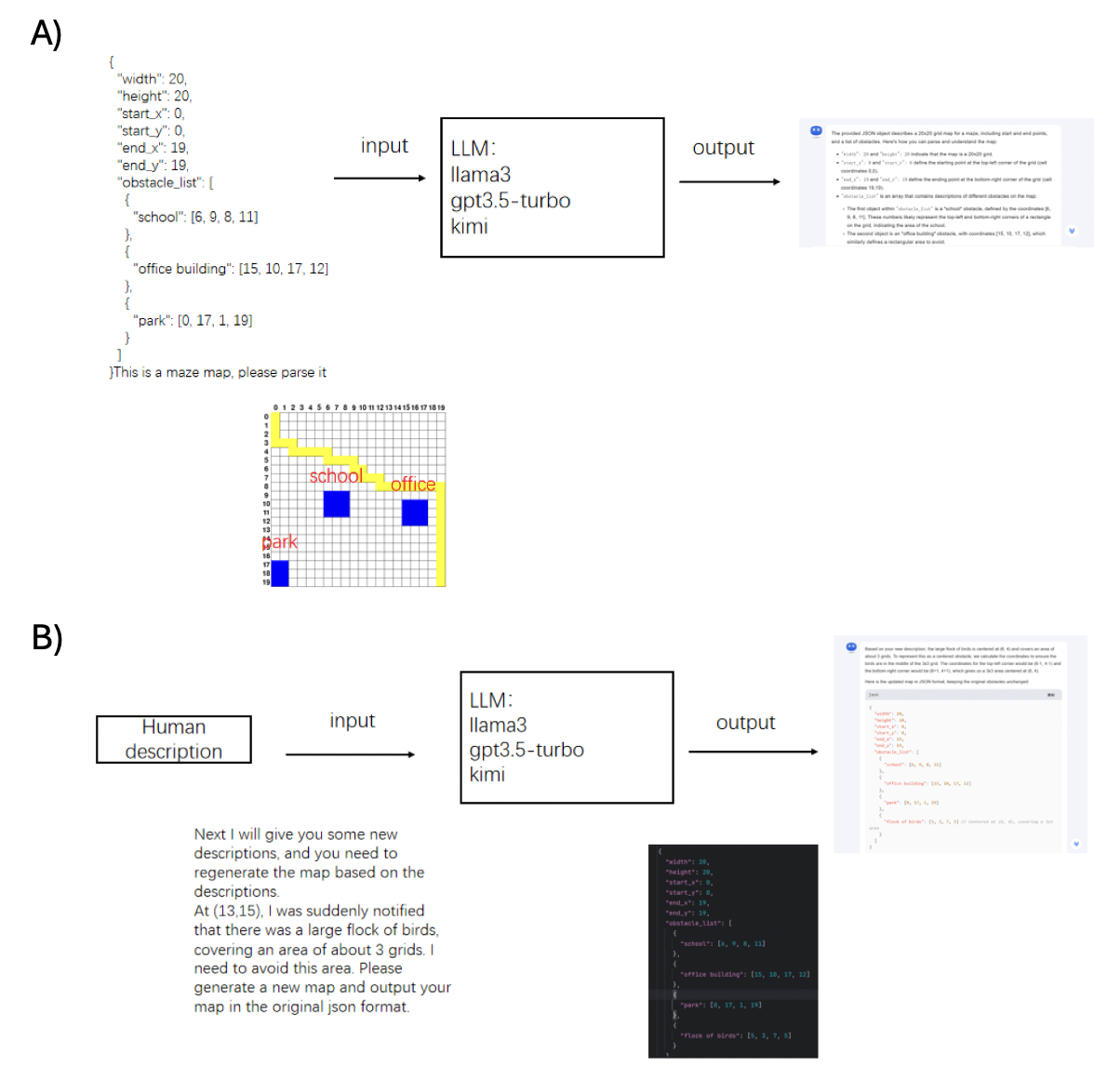}
\caption{A) Map parsing with descriptive inputs, B) Potential field adjustment with contextual inputs.}
\label{fig2}
\end{figure}

\begin{figure*}[!t]
\centering
\includegraphics[width=6.5in]{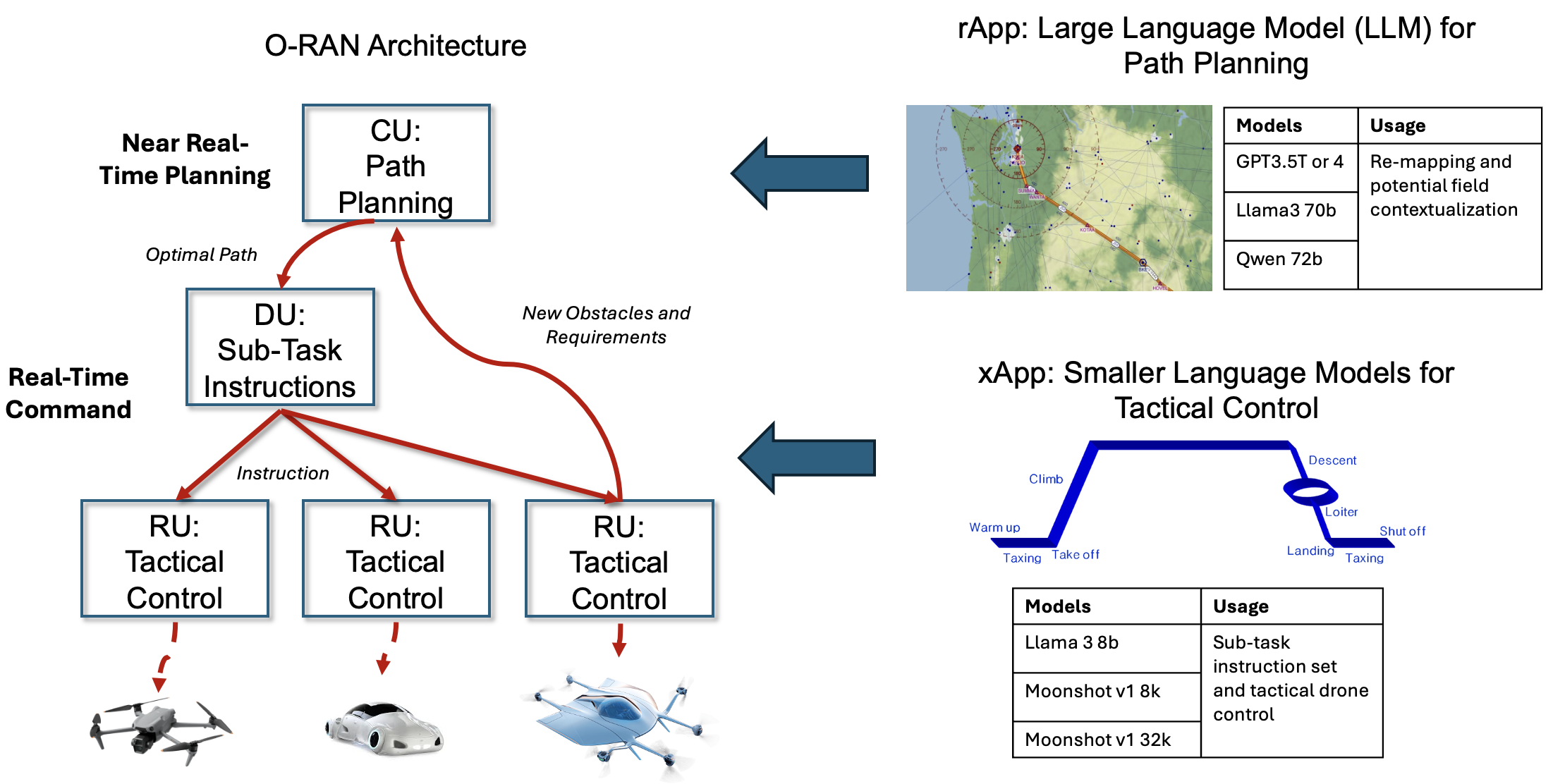}
\caption{Native-AI implemented inside ORAN architecture to achieve near real-time path planning using LLMs and real-time tactical control of sub-tasks using smaller language models.}
\label{fig3}
\end{figure*}

\subsection{LLM-informed Navigation: Path Planning Level}
For the LLM to adjust navigation algorithms (Fig.\ref{figs1}A-C), it is first necessary to inform the LLM of the format of the original map, allowing the LLM to understand and analyze the meaning of the JSON-formatted map. Subsequently, the changes in the map must be conveyed to the LLM in human language, and the LLM is required to comprehend the updated information about the map. It should then alter the map according to the construction method of the original map, outputting the new map in the JSON format of the original map. Finally, the map outputted by the LLM needs to be passed to path planning algorithms, such as the A* algorithm, to generate a new path.

\subsubsection{Map Information Parsing}
In this process, the LLM's input is map information in JSON format, and the output is its understanding of the map information. The understanding of the map information here is merely to verify whether the LLM has truly comprehended the JSON-formatted map and does not serve as subsequent input. However, it should be noted that the LLM has the capability to remember context. This part of the input is also saved in the LLM's memory, which means that in subsequent steps, the LLM can also know in what form the original map was constructed without additional input.

Parsing this map involves creating a 20x20 grid in a data structure see Fig.\ref{fig2}A, such as a two-dimensional array. Each cell of the grid would be initialized to represent open space. The obstacles would then be used to mark certain cells as obstacles, which should be avoided when calculating a path from the start to the end.
For a navigation algorithm, the grid would be scanned to identify the starting point and end-point. Then, the algorithm would consider the map's dimensions and the positions of the obstacles to determine a viable path while adhering to the no-fly zones. The specific logic for avoiding obstacles would depend on the algorithm used. 

\subsubsection{LLM-adjusted Potential Fields}
At this stage, the LLM needs to dynamically adjust the map based on the description in human language and output it in the original JSON map format, as shown in the Fig.\ref{fig2}B. Note that this stage must be conducted continuously with the previous process involving the LLM; otherwise, without the format information of the map in the LLM's memory, it cannot output the correct map format:
\begin{itemize}
    \item Example Input: "Next I will give you some new descriptions, and you need to regenerate the map based on the description. At (13, 15), I was suddenly notified that there was a large flock of birds, covering an area of about 3 grids. I need to avoid this area. Please generate a new map and output your map in the original json format."
    \item Output: “To incorporate the new description into the existing map and regenerate it while adhering to the original JSON format, you would add an entry for the bird flock to the (obstacle list). The area covered by the birds is approximately 3 grids, so you might define a rectangular area to represent this on the map.
\end{itemize}
This JSON structure maintains the original format and includes the new obstacle, allowing any navigation algorithm that uses this map to navigate around the bird flock as well.

\subsection{LLM-informed Navigation: Sub-Task Instruction Level}
For the LLM to create instruction sets of sub-task sequences (Fig.\ref{figs1}D), this is traditionally executed by a human operator following rules and regulations that include for example pre-flight check lists,...etc. 

Here, we use Moonshot LLM to create instruction sets, subject to mission requirements:
\begin{enumerate}
    \item Pre-Flight Check: Before takeoff, perform a pre-flight check to ensure the drone and its systems are operational.
    \item Takeoff: Initiate the takeoff sequence, lifting the drone off the ground to a safe altitude for the initial hover.
    \item Hover: Once airborne, hover at the starting point (0, 0) to stabilize and confirm the drone's readiness to proceed.
    \item Navigate to First Waypoint: Begin the flight by flying forward to the first waypoint, maintaining a steady altitude and watching for any obstacles.
    \item Sequential Waypoints: For each subsequent waypoint in the path [(0, 1), (0, 2), ..., (19, 19)], I would fly forward to next coordinate, perform turn on approach, and adjust altitude if needed.
    \item Avoid Obstacles: Continuously scan for obstacles and be prepared to adjust the flight path to avoid them while keeping the drone on course.
    \item **Hover Before Turns**: Before making a significant directional change, I might choose to hover momentarily to assess the surroundings and ensure a clear path.
    \item Final Approach: As the drone approaches the final waypoint (19, 19), I would slow down and prepare for landing.
    \item Land: Execute a safe landing sequence, bringing the drone down gently to the ground at the endpoint.
    \item Post-Flight: After landing, perform a post-flight check to ensure the drone is intact and ready for future flights.
\end{enumerate}

\section{Experimental Setup}

The experimental setup parameters can be found in Appendix A at end of paper. 

\subsection{Evaluation Criteria}

Evaluation of LLMs generally centers around their outputs; in this case it is to see if the code generated or modified is correct or not compared to human coder. There are a number of evaluation criteria for error categories, each with different error sources \cite{nature, 10.1145/3703155, yehuda-etal-2024-interrogatellm}. 

Through the aforementioned system setup and methodology, we check the output of the LLMs and the evaluation criteria include the following items:
\begin{enumerate}
    \item Can the map be correctly understood?
    \item Can the map be correctly modified for the lunch break scenario? Errors include two types: failure to correctly modify the map and incorrect map format output.
    \item Can the map be correctly modified for the emerging scenarios (e.g., bird flock)? Errors include two types: failure to correctly modify the map and incorrect map format output.
    \item Can the path planned by the algorithm be correctly understood and translated into flight instructions in human language? Errors include misunderstanding of the path and incorrect output in human language.
    \item Can the path be correctly understood and transformed into drone control commands? Errors include incorrect understanding of the path and incorrect drone action commands.
\end{enumerate}

\subsection{Comparison of Different LLMs in Path Planning}
In Table\ref{tab_size}, we can see that the smallest model sizes are Llama3:8b and possibly the Moonshot v1-8k model for short texts. These are more likely to be capable of edge deployment, whereas GPT and Qwen are definitely cloud deployment capabilities. We also note that sovereignty matters because drones and autonomous vehicles are critical national infrastructure entities and data transfer to clouds overseas may pose issues. 

In Table\ref{tab_benchmark1} and Table\ref{tab_benchmark2}, we see the performance of different LLMs for different map sizes and obstacle numbers. We can see that amongst the aforementioned 5 tests, larger models (e.g., those with over 70bn parameters) perform better - passing all 5 tests. Significantly smaller models have the potential for sovereign edge deployment, reducing communication latency and retaining data at the point of operation.

\begin{table}[ht]
    \centering
    \caption{LLM Model Size.}
    \label{tab_size}
    \begin{tabular}{lccc}
    \hline
    Model & Parameters & Context Tokens & Nation \\
    \hline
    GPT 3.5T    & 175Bn & 4k & USA \\
    GPT 4       & 1000Bn & 8-32k & USA \\
    Llama3:8b   & 8Bn & 8k & USA \\
    Llama3:70b  & 70Bn & 8k & USA \\
    Moonshot v1-8k & ? & 8k & China \\
    Moonshot v1-32k & ? & 32k & China  \\
    Qwen 72b    & 72Bn & 32k & China \\
    Cedille     & 6Bn  & 2k   & France \\
    \end{tabular}
\end{table}

\begin{table}[ht]
    \centering
    \caption{LLM Evaluation for Contextualized Path Planning for 3 Obstacles on $20\times20$ map: (1) Map Correctly Understood, (2) Map Correctly Modified, (3) Potential Field Correctly Modified, (4) Flight Sub-Task Instruction Generation, (5) Drone Control Command Generation.}
    \label{tab_benchmark1}
    \begin{tabular}{lccccc}
    \hline
    Model & 1 & 2 & 3 & 4 & 5 \\
    \hline
    GPT 3.5T & \checkmark & \checkmark & \checkmark & \checkmark & \checkmark \\
    GPT 4 & \checkmark & \checkmark & \checkmark & \checkmark & \checkmark \\
    Llama3:8b & \checkmark & \textbf{\textcolor{red}{Format Error}} & \checkmark & \checkmark & \checkmark \\
    Llama3:70b & \checkmark & \checkmark & \checkmark & \checkmark & \checkmark \\
    Moonshot v1-8k & \checkmark & \textbf{\textcolor{red}{Format Error}} & \checkmark & \checkmark & \checkmark \\
    Moonshot v1-32k & \checkmark & \checkmark & \checkmark & \checkmark & \checkmark \\
    Qwen 72b & \checkmark & \checkmark & \checkmark & \checkmark & \checkmark \\
    \end{tabular}
\end{table}

\begin{table}[ht]
    \centering
    \caption{LLM Evaluation for Contextualized Path Planning for 10 Obstacles on $40\times40$ map: (1) Map Correctly Understood, (2) Map Correctly Modified, (3) Potential Field Correctly Modified, (4) Flight Sub-Task Instruction Generation, (5) Drone Control Command Generation.}
    \label{tab_benchmark2}
    \begin{tabular}{lccccc}
    \hline
    Model & 1 & 2 & 3 & 4 & 5 \\
    \hline
    GPT 3.5T & \checkmark & \checkmark & \textbf{\textcolor{red}{Format}} & \checkmark & \checkmark \\
    GPT 4 & \checkmark & \checkmark & \checkmark & \checkmark & \checkmark \\
    Llama3:8b & \checkmark & \textbf{\textcolor{red}{Format }} & \textbf{\textcolor{red}{Format}} & \checkmark & \checkmark \\
    Llama3:70b & \checkmark & \checkmark & \checkmark & \checkmark & \checkmark \\
    Moonshot v1-8k & \checkmark & \textbf{\textcolor{red}{Format}} & \textbf{\textcolor{red}{Format}} & \checkmark & \checkmark \\
    Moonshot v1-32k & \checkmark & \checkmark & \checkmark & \checkmark & \checkmark \\
    Qwen 72b & \checkmark & \checkmark & \checkmark & \checkmark & \checkmark \\
    \end{tabular}
\end{table}

The following aspects are worth delving into:
1. Employing more complex and realistic simulation environments. Simulation environments similar to Airsim could be utilized, expanding the map into three-dimensional space, and constructing Airsim simulations based on local real-world maps. This would ensure a multitude of buildings, providing richer landmark structures for further drone navigation instructions under human language guidance.
2. Integrating human language instruction path execution into the drone, in conjunction with information captured by the drone's camera sensors, enables it to autonomously execute the planned path.

\subsection{Next Steps: Future 6G Architecture for Native LLM-Navigation}
As shown in Fig.\ref{fig3}, the future enabling computing-communication architecture will sit inside an Open Radio Access Network (ORAN) as part of 6G telecommunication vision. Here, native AI including LLMs will be embedded in the telecommunication GPUs, to serve both the telecommunication and its application end-user services. In this case, the LLMs will act as both real-time tactical command via (xApps) and near real-time path planning (rApps). These Apps are programmable into ORAN and act as a near operation computation capability.

We saw from Table\ref{tab_size} that some smaller LLMs are suitable for edge implementation, sitting at the edge Distributed Unit (DU) and Radio Unit (RU) architecture level, offering low latency command and control for autonomous platforms. Other larger LLMs are suitable for slow but accurate path planning, taking into account wide range of contextual requirements and obstacle data - sitting in the cloud Central Unit (CU).

\section{Conclusions and Future Work}

In this work, we analyzed the entire drone navigation process and found that traditional algorithms often fail to utilize information in human language to adjust maps in a timely manner when faced with dynamically changing environments; moreover, the output of the algorithms is based on mathematical model maps, which is not conducive to guiding drone pilots. Therefore, we introduced large language models (LLMs) as a bridge between the mathematical model world and the real human world. This approach can quickly and dynamically adjust maps based on instructions in human language, and on the other hand, they can quickly generate drone navigation instructions in human language based on the results of path planning algorithms. Furthermore, they can even directly generate flight control instructions for drone navigation. 

We conducted various experiments with multiple LLMs and different maps, and the results prove that our work is reliable. Our work not only has good practical significance but also proposes a general framework for LLM-assisted drone navigation tasks, thus allowing for further research within our framework. We also compared LLM performances and suggest architecturally how to deploy them for future 6G native AI architectures. 

Our future work will look to enhancing the LLM performance and will need to use Retrieval Augmented Generation (RAG) and inductive logic to achieve better inductive reasoning \cite{NIPS21}.

\qquad

\section{Appendix A - Experimental Setup}

Taking the LLM-assisted drone navigation task mentioned earlier as an example, the experiment includes the following steps:
\begin{enumerate}
    \item Input: width: 20, height: 20, $start_x$: 0, $start_y$: 0, $end_x$: 19, $end_y$: 19, obstacle list: school (6, 9, 8, 11), office building (15, 10, 17, 12), park (0, 17, 1, 19). This is a maze map - please analyze it.
    \item Input: Next, I will provide you with some new descriptions and you need to regenerate the map based on those descriptions.
    \item Input: On the basis of the original map, 1 PM is the break time at the school, so the area within two squares around it should be avoided. Please generate a new map and output it in the original JSON format.
    \item Input: On the basis of the original map, we have received a notice that there is a flock of birds on the left side of the school, and this area should be avoided. Please generate a new map and output it in the original JSON format.
    \item Input: according to the path planning algorithm, the resulting path is $[(0, 0), (0, 1), (0, 2), (0, 3), (1, 3), (2, 3), (2, 4), (3, 4)$, $(4, 4), (5, 4), (6, 4), (6, 5), (7, 5), (8, 5), (9, 5), (9, 6),$ $(10, 6), (10, 7), (11, 7), (12, 7), (12, 8), (13, 8), (14, 8)$, $(15, 8), (16, 8), (17, 8), (18, 8), (19, 8), (19, 9), (19, 10)$, $(19, 11), (19, 12), (19, 13), (19, 14), (19, 15), (19, 16)$, $(19, 17), (19, 18), (19, 19)]$. Assuming you are a professional drone pilot, you need to translate the output of the path planning algorithm into a navigation language that you can understand. How would you do this?
    \item Input: Assuming you are a professional drone pilot and need to actually execute the drone flight according to the method mentioned earlier, here is how you would control the drone, assuming the drone's movements include (takeoff, landing, hovering, flying forward, and turning left, right, and around), and noting that you need to consider takeoff and landing at the start and end point.
\end{enumerate}

\bibliographystyle{IEEEtran}
\bibliography{main.bib}

\end{document}